\documentclass[10pt,twocolumn,letterpaper]{article}

\usepackage{iccv}              %

\newcommand{\cmark}{\ding{51}}%
\newcommand{\xmark}{\ding{55}}%

\usepackage{makecell}
\usepackage{multirow}
\usepackage{colortbl}
\usepackage{textcomp} 
\usepackage{indentfirst}
\usepackage{pifont}

\definecolor{nbarrier}{RGB}{255, 120, 50}
\definecolor{nbicycle}{RGB}{255, 192, 203}
\definecolor{nbus}{RGB}{255, 255, 0}
\definecolor{ncar}{RGB}{0, 150, 245}
\definecolor{nconstruct}{RGB}{0, 255, 255}
\definecolor{nmotor}{RGB}{200, 180, 0}
\definecolor{npedestrian}{RGB}{255, 0, 0}
\definecolor{ntraffic}{RGB}{255, 240, 150}
\definecolor{ntrailer}{RGB}{135, 60, 0}
\definecolor{ntruck}{RGB}{160, 32, 240}
\definecolor{ndriveable}{RGB}{255, 0, 255}
\definecolor{nother}{RGB}{139, 137, 137}
\definecolor{nsidewalk}{RGB}{75, 0, 75}
\definecolor{nterrain}{RGB}{150, 240, 80}
\definecolor{nmanmade}{RGB}{213, 213, 213}
\definecolor{nvegetation}{RGB}{0, 175, 0}

\definecolor{nvcolor}{RGB}{119,185,0}
\definecolor{roadcolor}{RGB}{234,51,246}
\definecolor{sidewalkcolor}{RGB}{68,8,72}
\definecolor{parkingcolor}{RGB}{241,156,249}
\definecolor{othergroundcolor}{RGB}{160,32,76}
\definecolor{buildingcolor}{RGB}{246,202,69}
\definecolor{carcolor}{RGB}{111,149,238}
\definecolor{truckcolor}{RGB}{74,32,172}
\definecolor{bicyclecolor}{RGB}{136,227,242}
\definecolor{motorcyclecolor}{RGB}{37,59,146}
\definecolor{othervehiclecolor}{RGB}{96,81,242}
\definecolor{vegetationcolor}{RGB}{79, 173, 50}
\definecolor{trunkcolor}{RGB}{126, 65, 22}
\definecolor{terraincolor}{RGB}{171, 238, 105}
\definecolor{personcolor}{RGB}{234, 60, 49}
\definecolor{bicyclistcolor}{RGB}{234, 66, 195}
\definecolor{motorcyclistcolor}{RGB}{138, 42, 90}
\definecolor{fencecolor}{RGB}{238, 128, 69}
\definecolor{polecolor}{RGB}{252, 241, 161}
\definecolor{trafficsigncolor}{RGB}{233, 51, 35}
\definecolor{other-struct.color}{RGB}{255, 150, 0}
\definecolor{other-objectcolor}{RGB}{50, 255, 255}
\definecolor{lane-markingcolor}{RGB}{150, 255, 170}
\definecolor{color1}{RGB}{176, 36, 24}
\definecolor{color2}{RGB}{0, 176, 80}
\definecolor{color3}{RGB}{0, 0, 200}

\definecolor{iccvblue}{rgb}{0.21,0.49,0.74}
\usepackage[pagebackref,breaklinks,colorlinks,allcolors=iccvblue]{hyperref}

\def\paperID{12861} %
\def\confName{ICCV}
\def\confYear{2025}

\newcommand{\ourmethod}{\textsc{S2GO}\xspace}

\title{\ourmethod: Streaming Sparse Gaussian Occupancy Prediction}

\author{
Jinhyung Park\textsuperscript{1,2\thanks{Work done during internship at Applied Intuition}} \hspace{0.7em}
Yihan Hu\textsuperscript{1} \hspace{0.7em}
Chensheng Peng\textsuperscript{1,3*} \hspace{0.7em}
Wenzhao Zheng\textsuperscript{3} \hspace{0.7em}
Kris Kitani\textsuperscript{2} \hspace{0.7em} 
Wei Zhan\textsuperscript{1,3\thanks{Correspondence: \href{wei.zhan@applied.co}{wei.zhan@applied.co}}} \hspace{0.7em} \\
\textsuperscript{1}Applied Intuition \hspace{0.7em}
\textsuperscript{2}Carnegie Mellon University \hspace{0.7em}
\textsuperscript{3}University of California, Berkeley
}

\begin{document}
\maketitle
\begin{abstract}
Despite the demonstrated efficiency and performance of sparse query-based representations for perception, state-of-the-art 3D occupancy prediction methods still rely on voxel-based or dense Gaussian-based 3D representations. However, dense representations are slow, and they lack flexibility in capturing the temporal dynamics of driving scenes. Distinct from prior work, we instead summarize the scene into a compact set of 3D queries which are propagated through time in an online, streaming fashion. These queries are then decoded into semantic Gaussians at each timestep. We couple our framework with a denoising rendering objective to guide the queries and their constituent Gaussians in effectively capturing scene geometry. Owing to its efficient, query-based representation, \ourmethod achieves state-of-the-art performance on the nuScenes and KITTI occupancy benchmarks, outperforming prior art (e.g., GaussianWorld) by \textbf{1.5 IoU} with \textbf{5.9x faster} inference.

\end{abstract}

\section{Introduction}

Vision-centric autonomous systems provide a more cost-effective and scalable alternative to LiDAR-based solutions~\cite{wayve2024cvprw, mobileye2024ces, tesla2022aiday, zhang2024vision}, yet they struggle with the absence of dense 3D geometry priors—an obstacle to achieving beyond Level 3 autonomy. To address this gap, 3D occupancy semantic prediction has emerged as a powerful complement to conventional sparse 3D perception tasks like bounding box detection~\cite{philion2020lift, yin2021center, petr, streampetr, bevformer, bevdepth} or vectorized mapping~\cite{li2021hdmapnet, liao2022maptr, liao2024maptrv2, chen2024maptracker, yuan2024streammapnet}, because it captures a richer and more comprehensive view of unknown and arbitrarily shaped objects, thereby improving safety.

Recent 3D occupancy methods often rely on regular grids~\cite{monoscene,bevformer,occformer,tpvformer,liu2024fully} or dense Gaussians \cite{gaussianformer,zuo2024gaussianworld,zheng2024gaussianad}.
Although these methods capture high-fidelity details, they are slow and inflexible when integrating long-term historical context,  limiting both static infrastructure localization as well as dynamic actor modeling. Existing grid-based approaches reduce redundancy by warping or projecting features from previous frames~\cite{bevformer, bevdet4d, fb-occ, liu2023sparsebev, liu2024fully}, but suffer from unnecessary computation in unoccupied regions and artifacts introduced by dense grids. Meanwhile, recent Gaussian-based techniques~\cite{gaussianformer, huang2024probabilistic, zheng2024gaussianad, zuo2024gaussianworld} show promise by focusing computation on occupied regions. However, they rely on tens of thousands of Gaussians (25.6k~$\sim$~144k) and use local sparse convolutions because global modeling becomes computationally prohibitive.

To address the inefficiencies of voxel-based and dense Gaussian-based methods in streaming perception, we propose to use \textit{sparse 3D queries} to summarize and propagate the \textit{dense 3D world} over time. More specifically, our method (\textbf{\ourmethod}) maintains a queue of past sparse 3D queries, refines the current set of queries using both previous queries and current image observations, and then predicts 3D occupancy by decoding the current queries into a denser set of semantic Gaussians. This online framework enables efficient propagation and global feature interaction among a sparser set of 3D queries ($\sim$1k) while retaining the high fidelity of Gaussian-based representations.

Query-based perception has demonstrated its effectiveness in sparse object detection \cite{carion2020detr,wang2020detr3d,streampetr}, but employing sparse queries for dense, high-fidelity occupancy prediction presents several challenges. \textbf{First}, object detectors typically employ hundreds to thousands of queries, which far outnumber the target objects (approximately 30 per scene), allowing for explicit one-to-one Hungarian Matching. In contrast, 3D occupancy estimation must cover the entire scene, making the mapping from sparse queries to dense semantic Gaussians inherently ambiguous. \textbf{Second}, in voxel-aligned occupancy prediction, fixed voxels simply perform classification at their predetermined locations. By comparison, query-based approaches require that queries first move to regions of interest before classifying.
This creates a chicken-and-egg problem: for instance, if a query lies between a car and the road, it is unclear whether it should shift toward the car or the road, as the correct target location depends on the query’s intended class. 
\textbf{Third}, while dense Gaussian methods mitigate this ambiguity through extensive spatial coverage, increasing the sparsity of the representation for efficiency exacerbates the difficulty of aligning queries accurately with occupied regions.

Some methods utilize grid- or voxel-based sparse queries for 3D occupancy prediction~\cite{liu2024fully,voxformer}, inherently limiting their effective use of long-term temporal information. To fully unlock the streaming potential of query-based occupancy prediction, we introduce a pre-training phase that trains the network to capture 3D scene geometry before the semantic occupancy stage. During pre-training, query locations are initialized with noised, evenly sampled LiDAR points, and the network is trained to recover 3D geometry through a denoising objective. To capture fine-grained local shape, decoded Gaussians are rendered from the current and neighboring views and supervised accordingly. The network also predicts a velocity for each query to model dynamic objects. 
This pre-training addresses the aforementioned challenges of using sparse queries by 1) supervising queries and their decoded Gaussians to model local scene structure, 2) training queries to self-organize to evenly cover the scene, and 3) supervising queries explicitly to move from empty space to occupied regions. Following this pre-training phase, during the semantic occupancy prediction stage -- when LiDAR data is no longer used and queries are randomly initialized throughout the 3D scene -- the network uses its pre-trained knowledge to precisely reposition the queries and decode to Gaussians to capture overall dense 3D structure.

Our contributions are summarized as follows. 
\begin{itemize} 
    \item We propose \textbf{\ourmethod}, an efficient and novel framework for 3D semantic occupancy prediction using sparse 3D queries. Our online, streaming approach effectively captures long-term historical context. 
    \item To address the challenge of making \textit{dense} predictions from a \textit{sparse} representation, we introduce a geometry denoising pre-training phase. This enables sparse 3D queries to move through empty space in order to reach and cover occupied regions while self-organizing to capture dense 3D structure. 
    \item We evaluate our pipeline on the nuScenes and KITTI benchmarks and achieve state-of-the-art performance and inference speed. Notably, our lightweight model improves over prior art (e.g. GaussianWorld) with 5.9$\times$ faster inference, achieving real-time inference on a single 4090 (26 FPS). 
\end{itemize}

\section{Related Work}
\label{sec:related-work}

\textbf{3D Occupancy Prediction} is increasingly crucial for vision-centric systems due to limited geometric priors inherent in purely vision-based methods. This task provides dense, volumetric representations of the environment, significantly enhancing semantic understanding and improving safety in decision-making, effectively complementing LiDAR. Recent camera-based benchmarks~\cite{wei2023surroundocc, tian2023occ3d, liao2022kitti}, featuring detailed annotations created through offboard techniques, have driven substantial progress in vision-based occupancy modeling research.

Building upon these benchmarks, existing methods~\cite{cao2022monoscene, li2023fb, huang2023tri, zhang2023occformer, ye2024cvt, zheng2024occworld, scene_as_occ} typically employ dense BEV or voxel-based representations, but such structures hinder real-time processing efficiency and scalability. Sparse-voxel approaches~\cite{liu2024fully, li2023voxformer, wang2024panoocc} enhance efficiency by introducing sparse representations, yet encounter challenges such as complex temporal modeling and increased overhead in temporal integration due to their grid-based nature.

Recently, Gaussian-based representations~\cite{kerbl20233d, peng2024desire, yang2024unipad, xu2024gaussianpretrain} have emerged in autonomous driving due to their strong 3D and semantic representational capabilities. Methods such as~\cite{huang2024gaussianformer, zheng2024gaussianad, zuo2024gaussianworld} exploit probabilistic semantic Gaussians for 3D occupancy modeling, but they typically require large numbers of Gaussians, posing challenges for real-time performance and efficient temporal fusion. Also related is OSP \cite{occaspoints}, which represents the scene as a set of points. While flexible, sparse points cover a narrower region of the scene compared to Gaussians, and OSP requires grid-aligned point sampling to make voxel-aligned predictions.

\vspace{5pt}
\textbf{Query-based Representations.} Since DETR~\cite{carion2020detr}, query-based methods have rapidly advanced, demonstrating effectiveness in tasks like detection, mapping, and tracking. DETR3D~\cite{luo2022detr4d} efficiently extends 2D queries into 3D for detection, while StreamPETR~\cite{wang2023exploring} fuses temporal information in a streaming fashion. MapTR~\cite{liao2022maptr, liao2024maptrv2} leverages structured Transformers for HD map generation, and MapTracker~\cite{chen2024maptracker, yuan2024streammapnet} reframes the mapping task with object tracking. Sparse4D~\cite{lin2023sparse4dv3} integrates detection and tracking into a unified, end-to-end framework. However, object-centric query methods remain underutilized for dense reconstruction tasks like occupancy prediction. We bridge this gap by introducing Gaussian queries, establishing a streamlined, query-based framework for efficient 3D semantic occupancy prediction.

\begin{figure*}[t]
  \centering
  \vspace{-0.2in}
  \includegraphics[width=\linewidth]{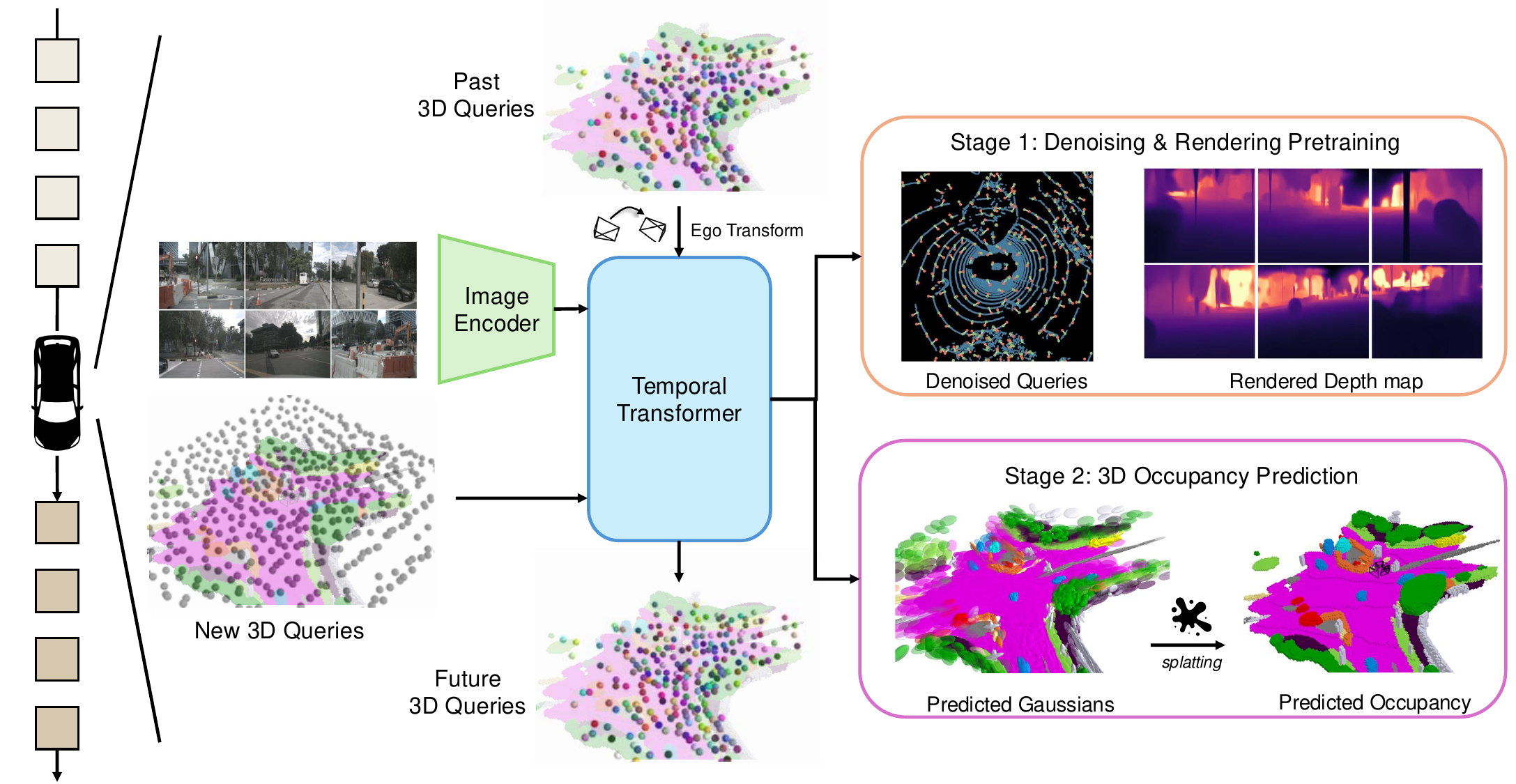}
  \vspace{-0.3in}
  \caption{\textbf{Overall framework of \ourmethod for streaming perception.} At each timestep, our method refines new 3D queries using current image observations and a queue of past queries. These queries are decoded into a set of fine-grained Gaussians, and a portion of the queries are propagated to future timesteps in a streaming fashion. In Stage 1, this query refinement and Gaussian prediction pipeline is pre-trained to effectively model the 3D scene using query denoising and rendering pre-training. In Stage 2, the predicted Gaussians are splatted to voxels for training 3D occupancy prediction. }
  \label{fig:main_diagram}
  \vspace{-0.15in}
\end{figure*}

\section{\ourmethod}

\subsection{Preliminary: Gaussian Occupancy Prediction}\label{method:prelim_gauss}
GaussianFormer \cite{gaussianformer} and follow-up work \cite{huang2024probabilistic,zheng2024gaussianad,zuo2024gaussianworld} propose to represent the driving scene as a set of 3D Gaussian primitives $\mathcal{G}=\{\mathbf{G}_i\}_{i=1}^{P}$, with each semantic Gaussian $\mathbf{G}_i$ specified by its position $\mathbf{x}_i \in \mathbb{R}^3$, rotation $\mathbf{r}_i \in \mathbb{R}^4$, scale $\mathbf{s}_i \in \mathbb{R}^3$, opacity $a_i \in \mathbb{R}$ and class distribution $\mathbf{c}_i \in \mathbb{R}^C$, where $C$ is the number of foreground classes. Given this set, GaussianFormer-2 \cite{huang2024probabilistic} predicts the semantic occupancy of a voxel coordinate $\mathbf{x} \in \mathbb{R}^3$ by first predicting binary occupancy probability and then expressing the class distribution of occupied regions as a mixture of nearby Gaussians. More specifically, the occupancy probability $\alpha(\textbf{x}) \in \mathbb{R}$ is modeled as the probability that $\textbf{x}$ is occupied by at least one of $P$ nearby Gaussians: 
\vspace{-1em}
\begin{equation}\label{eq:fg_bg}
    \alpha(\mathbf{x}) = 1 - \prod_{i=1}^{P}\big(1 - \alpha(\mathbf{x};\mathbf{G}_i)\big)
    \vspace{-.5em}
\end{equation}
where $\alpha(\mathbf{x};\mathbf{G}_i)$ is the probability that $x$ is occupied by $\mathbf{G}_i$:
\vspace{-1em}
\begin{equation}\label{eq:gaussian_prob}
    \alpha(\mathbf{x};\mathbf{G}) = {\rm{exp}}\big(-\frac{1}{2}(\mathbf{x}-\mathbf{m})^{\rm T} \mathbf{\Sigma}^{-1} (\mathbf{x}-\mathbf{m})\big)
\end{equation}
\vspace{-1em}
\begin{equation}
    \mathbf{\Sigma} = \mathbf{R}\mathbf{S}\mathbf{S}^T\mathbf{R}^T, \quad \mathbf{S} = {\rm{diag}}(\mathbf{s}), \quad \mathbf{R} = {\rm{q2r}}(\mathbf{r})
\end{equation}
Further, the foreground class distribution $\mathbf{e}(\mathbf{x};\mathcal{G}) \in \mathbb{R}^C$ is expressed as a mixture of Gaussians weighted by opacity $a$:
\vspace{-1em}
\begin{equation}\label{eq:mix}
\begin{aligned}
    \mathbf{e}(\mathbf{x};\mathcal{G}) &= \sum_{i=1}^{P} p(\mathbf{G}_i|\mathbf{x})\Tilde{\mathbf{c}}_i 
    = \frac{\sum_{i=1}^{P}p(\mathbf{x}|\mathbf{G}_i)a_i\Tilde{\mathbf{c}}_i}{\sum_{j=1}^{P}p(\mathbf{x}|\mathbf{G}_j)a_j},
\end{aligned}
\end{equation}
\vspace{-.5em}
\begin{small}
\begin{equation}
    p(\mathbf{x}|\mathbf{G}_i) = \frac{1}{(2\pi)^{\frac{3}{2}}|\mathbf{\Sigma}|^{\frac{1}{2}}}{\rm{exp}}\big(-\frac{1}{2}(\mathbf{x}-\mathbf{m})^{\rm T} \mathbf{\Sigma}^{-1} (\mathbf{x}-\mathbf{m})\big)
\end{equation}
\end{small}
Finally, the joint semantic occupancy distribution over foreground classes and the empty background is written as $[\alpha(\mathbf{x})\cdot\mathbf{e}(\mathbf{x};\mathcal{G}); 1-\alpha(\mathbf{x})] \in \mathbb{R}^{(C+1)}$. We refer the reader to prior work \cite{gaussianformer,huang2024probabilistic} for additional details.

\begin{figure*}[t]
  \centering
  \vspace{-0.2in}
  \includegraphics[width=\linewidth]{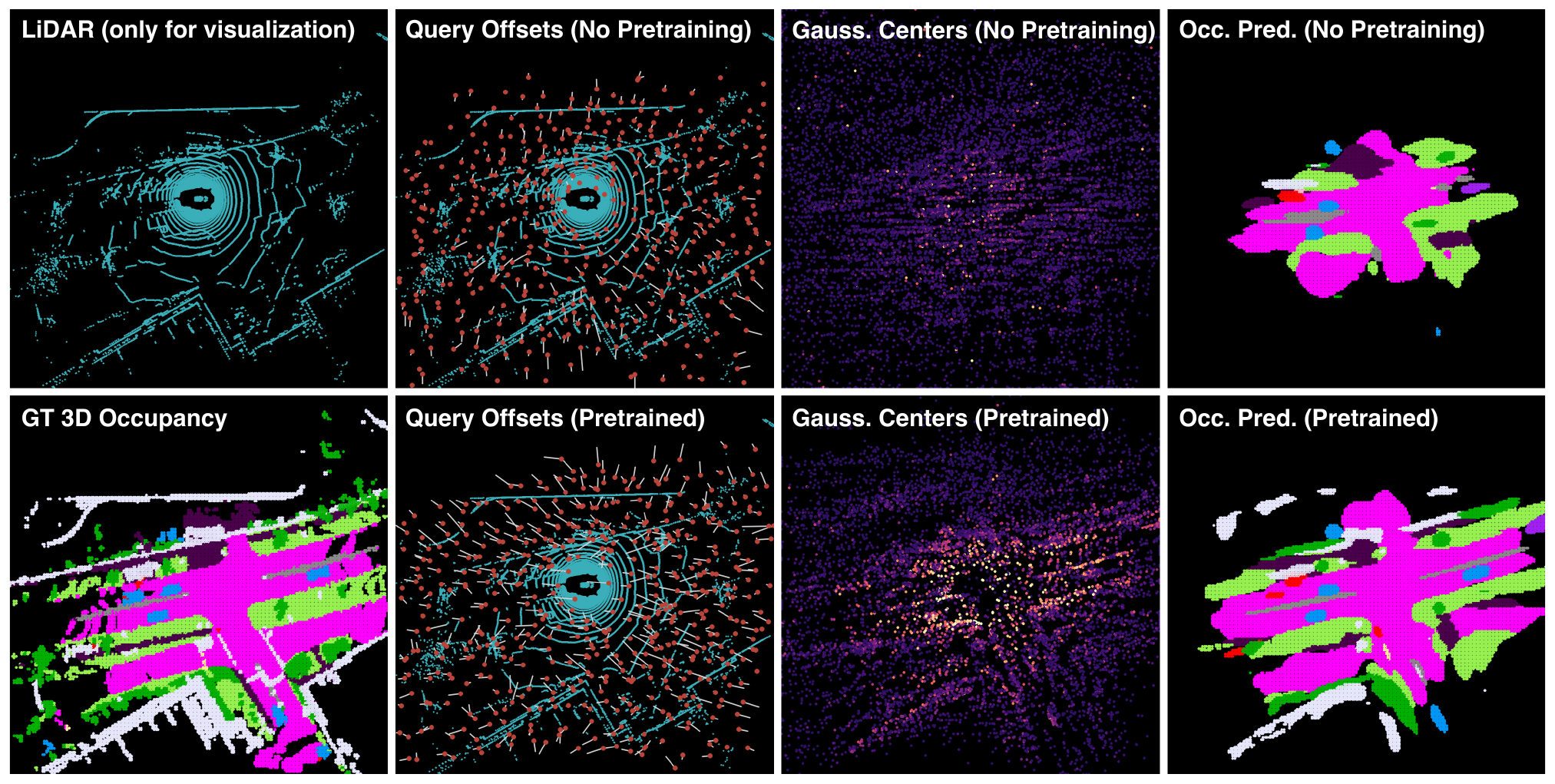}
  \vspace{-0.2in}
    \caption{\textbf{Impact of denoising pre-training on occupancy prediction.} We visualize query offsets (column 2), Gaussian centers (column 3) colored by opacity, and occupancy predictions (column 4) for \ourmethod~with and without pre-training. Without pre-training, queries remain largely stagnant, and Gaussians fail to capture 3D structures. In contrast, pre-training with rendering and denoising allows queries to move towards occupied regions—particularly visible for {\color[RGB]{128,128,128}\textbf{walls}} and {\color[RGB]{  0, 150, 245}\textbf{cars}}—while Gaussians self-organize to better represent the scene, significantly improving occupancy prediction. %
    }
  \label{fig:denoise_motivation}
\end{figure*}

\subsection{Architecture}\label{method:arch}
Our framework shown in Figure \ref{fig:main_diagram} is inspired by streaming query-based object detection methods \cite{carion2020detr,wang2020detr3d,streampetr,sparse4d,yuan2024streammapnet}. We keep a queue of past sparse 3D queries, update the current queries based on historical queries and current images, and predict a detailed set of 3D Gaussians. 

More specifically, at each timestep $t$, we represent the scene with a set of sparse 3D queries $\mathcal{Q}_t = \{\mathbf{q}_t^i\}_{i=1}^K$ with associated 3D locations $\{\mathbf{p}_t^i\}_{i=1}^K \subset \mathbb{R}^3$, where $K$ is the number of queries. These queries are refined using a queue of past queries $\bar{\mathcal{Q}}_t$ and the 2D features $\mathcal{F}_t = \text{CNN}(I_t)$ from the RGB images of that timestep $I_t \in \mathbb{R}^{N \times H\times W\times 3}$, where $N$ is the number of cameras.

Each query predicts a position offset $\mathbf{o}^i$, opacity $a^i$, and velocity $\mathbf{v}^i$, alongside attributes for a set of finer Gaussians. Relaxing the timestep $t$ subscript on Gaussians for clarity, the derived Gaussians are written as:
\begin{equation}
\mathcal{G}_t = \{\{(\mathbf{p}^i + \mathbf{o}^i + \mathbf{o}_j^i, \mathbf{v}^i, \mathbf{r}_j^i, \mathbf{s}_j^i, a^i \cdot a_j^i)\}_{j=1}^J\}_{i=1}^K
\end{equation}
where $J$ is the number of Gaussians per query. Each Gaussian has a 3D position $\mathbf{p}^i + \mathbf{o}^i + \mathbf{o}_j^i$ combining the query position, query offset, and its own offset $\mathbf{o}_j^i$, a velocity $\mathbf{v}^i$ inherited from its parent query, a rotation $\mathbf{r}_j^i$, a scale $\mathbf{s}_j^i$, and an opacity $a^i \cdot a_j^i$ where the query opacity modulates the Gaussian-specific opacity $a_j^i$. This hierarchical decomposition allows each query to anchor a spatial region, while the finer Gaussians capture local structure within that region.

 Our framework for efficiently extracting 3D Gaussians from image observations is consistent across both the denoising pretraining and occupancy prediction tasks. The primary distinction lies in the additional attributes predicted by each Gaussian: during pretraining, each Gaussian independently predicts its own color, whereas, during occupancy prediction, Gaussians derived from the same query collectively share a semantic class label. This shared semantic class ensures consistency among Gaussians originating from a single query.

\subsection{Stage 1: 3D Geometry Denoising}\label{method:stage1}
\subsubsection{Motivation}
While \ourmethod can directly be trained for occupancy prediction, the resulting performance is suboptimal. The queries and their Gaussians are unable to move effectively to occupied locations to capture fine details -- they instead coarsely model nearby regions as shown in Figure \ref{fig:denoise_motivation}. This stems from the weak and ambiguous supervision that queries and Gaussians receive from occupancy labels. 

This limitation arises from two interconnected factors: \textbf{First}, unlike in GaussianFormer where each Gaussian is refined individually, in our sparse query-based framework, each query moves $J$ Gaussians as a group before individual Gaussians locally branch out. As any perturbations to query location propagate to its constituent Gaussians, aligning the query precisely with scene geometry before predicting Gaussian offsets is critical. However, 3D occupancy prediction lacks a clear assignment between parts of the scene and individual queries -- with multiple nearby scene elements, the lack of clear-cut supervision causes query refinements to be noisy. \textbf{Second}, this ambiguity is exacerbated by the inherent \textit{locality} of the Gaussian-to-voxel splatting operation in Section \ref{method:prelim_gauss}. As Gaussians are each locally pulled to different scene elements -- suboptimal local minima \cite{kerbl20233d,charatan2024pixelsplat} -- their corresponding queries are similarly stuck in suboptimal locations, unable to properly cover the scene. 

\subsubsection{Denoising and Rendering Framework}
To explicitly supervise query movement and train Gaussians to model 3D geometry around their queries, we introduce a denoising and rendering framework for pre-training \ourmethod. The model functions as described in Section \ref{method:arch}, but in this stage, we initialize current query locations ${\mathbf{p}^i}$ at noised LiDAR points at that timestep. Relaxing the $t$ subscript, given 3D points $\mathbf{pts} \in \mathbb{R}^{M\times 3}$, we set
\begin{equation}
    \{\mathbf{p}^i\}_{i=0}^K = \text{FPS}_K(\mathbf{pts}) + \epsilon
\end{equation}
where $M$ is the \# of LiDAR points, $\epsilon \sim \text{U}(-e, e)^{K\times3}$, $\text{FPS}_K$ applies Furthest-Point-Sampling (FPS) to yield $K$ points, and $\text{U}(-e, e)$ is the continuous uniform distribution with $e$ as a hyperparameter. Starting at these noised positions, the model predicts query offsets $\{\mathbf{o}^i_t\}_{i=1}^K$ and derived Gaussians $\mathcal{G}_t$ for the current scene.
\subsubsection{Training Objectives}
We then supervise these outputs with the loss function:
{\small
\begin{equation}
    \begin{aligned}
        \mathcal{L} =& \lambda_1 \sum_{i=1}^K||\text{FPS}_K(\mathbf{pts_t}) - (p^i_t + o^i_t)|| \\
        &+ \lambda_2\mathcal{L}_{depth}(\mathcal{G}, D) + \lambda_3\mathcal{L}_{rgb}(\mathcal{G}, I)
    \end{aligned}
\end{equation}
}
The first term is the denoising objective, training the network to self-organize the queries to cover 3D structure. Then, $\mathcal{L}_{depth}$ and $\mathcal{L}_{rgb}$ render depth maps and RGB images from the Gaussians and supervise them with LiDAR projected depth maps $D_t$ and image observations. This explicitly trains Gaussians to represent detailed scene structure around the aligned queries. 
Notably, the rendering supervision is done on current and neighboring keyframes (+/- 0.5s) by moving the Gaussians with predicted velocities $v$ and accounting for ego-motion. This further improves final 3D occupancy performance.
Altogether, this denoising and rendering stage provides \ourmethod with a strong prior for sparse queries and Gaussians to effectively model the 3D scene geometry.

\begin{table*}[t] %
    \small
    \setlength{\tabcolsep}{0.005\linewidth}  
    \renewcommand\arraystretch{1.05}
    \centering
        \vspace{-3mm}  
    \resizebox{\textwidth}{!}{
    \begin{tabular}{l|c c | c c c c c c c c c c c c c c c c | c}
        \toprule
        Method
        & IoU
        & mIoU
        & \rotatebox{90}{\textcolor{nbarrier}{$\blacksquare$} barrier}
        & \rotatebox{90}{\textcolor{nbicycle}{$\blacksquare$} bicycle}
        & \rotatebox{90}{\textcolor{nbus}{$\blacksquare$} bus}
        & \rotatebox{90}{\textcolor{ncar}{$\blacksquare$} car}
        & \rotatebox{90}{\textcolor{nconstruct}{$\blacksquare$} const. veh.}
        & \rotatebox{90}{\textcolor{nmotor}{$\blacksquare$} motorcycle}
        & \rotatebox{90}{\textcolor{npedestrian}{$\blacksquare$} pedestrian}
        & \rotatebox{90}{\textcolor{ntraffic}{$\blacksquare$} traffic cone}
        & \rotatebox{90}{\textcolor{ntrailer}{$\blacksquare$} trailer}
        & \rotatebox{90}{\textcolor{ntruck}{$\blacksquare$} truck}
        & \rotatebox{90}{\textcolor{ndriveable}{$\blacksquare$} drive. suf.}
        & \rotatebox{90}{\textcolor{nother}{$\blacksquare$} other flat}
        & \rotatebox{90}{\textcolor{nsidewalk}{$\blacksquare$} sidewalk}
        & \rotatebox{90}{\textcolor{nterrain}{$\blacksquare$} terrain}
        & \rotatebox{90}{\textcolor{nmanmade}{$\blacksquare$} manmade}
        & \rotatebox{90}{\textcolor{nvegetation}{$\blacksquare$} vegetation}
        & FPS
        \\
        \midrule
        MonoScene~\cite{cao2022monoscene} & 23.96 & 7.31 & 4.03 &	0.35& 8.00& 8.04&	2.90& 0.28& 1.16&	0.67&	4.01& 4.35&	27.72&	5.20& 15.13&	11.29&	9.03&	14.86 & -\\
        
        Atlas~\cite{murez2020atlas} & 28.66 & 15.00 & 10.64&	5.68&	19.66& 24.94& 8.90&	8.84&	6.47& 3.28&	10.42&	16.21&	34.86&	15.46&	21.89&	20.95&	11.21&	20.54 & -\\
        
        BEVFormer~\cite{li2022bevformer} & 30.50 & 16.75 & 14.22 &	6.58 & 23.46 & 28.28& 8.66 &10.77& 6.64& 4.05& 11.20&	17.78 & 37.28 & 18.00 & 22.88 & 22.17 & {13.80} &	{22.21} & 3.3\\

        TPVFormer~\cite{huang2023tri}  & {30.86} & 17.10 & 15.96&	 5.31& 23.86	& 27.32 & 9.79 & 8.74 & 7.09 & 5.20& 10.97 & 19.22 & {38.87} & {21.25} & {24.26} & {23.15} & 11.73 & 20.81 & 2.9\\

        OccFormer~\cite{zhang2023occformer} & {31.39} & {19.03} & {18.65} & {10.41} & {23.92} & {30.29} & {10.31} & {14.19} & {13.59} & {10.13} & {12.49} & {20.77} & {38.78} & 19.79 & 24.19 & 22.21 & {13.48} & {21.35} & -\\
        
        SurroundOcc~\cite{wei2023surroundocc} & {31.49} & {20.30}  & {20.59} & {11.68} & {28.06} & {30.86} & {10.70} & {15.14} & {14.09} & {12.06} & {14.38} & {22.26} & 37.29 & {23.70} & {24.49} & {22.77} & {14.89} & {21.86} & 3.3 \\

        GaussianFormer~\cite{huang2024gaussianformer} & 29.83 & {19.10} & {19.52} & {11.26} & {26.11} & {29.78} & {10.47} & {13.83} & {12.58} & {8.67} & {12.74} & {21.57} & {39.63} & {23.28} & {24.46} & {22.99} & 9.59 & 19.12 & 2.7\\

        GaussianFormer-2~\cite{huang2024probabilistic} & {31.74} & {20.82} & {21.39} & {13.44} & {28.49} & 30.82 & 10.92 & {15.84} & 13.55 & 10.53 & 14.04 & {22.92} & {40.61} & {24.36} & {26.08} & {24.27} & 13.83 & 21.98  & 2.8 \\

        {GaussianWorld*}~\cite{zuo2024gaussianworld} & 32.77  & 21.79 & 21.61 & 13.30 & 27.28 & 31.21 
                   & 13.89 & 16.91 & 13.28 & 11.77 & 14.82 & 23.66 
                   & 41.91 & 24.31 & 28.35 & 26.32 & 15.67 & 24.54 & 4.4\\

        \midrule
        S2GO-Small & 34.27 & 22.11 & 20.80 & 13.08 & 27.46 & 30.25 & 14.50 & 16.50 & 11.72 & 10.92 & 13.54 & 23.26 & 46.29 & 29.19 & 29.72 & 28.44 & 13.02 & 25.05 & \textbf{26.1} \\
        S2GO-Base & \textbf{35.46} & \textbf{22.72} & 21.93 & 13.36 & 27.47 & 32.08 & 14.86 & 15.31 & 12.91 & 11.79 & 13.42 & 23.98 & 46.85 & 29.14 & 30.30 & 29.05 & 14.69 & 26.40 & 19.6 \\
        \bottomrule
    \end{tabular}}
    \vspace{-.5em}
    \caption{\textbf{3D occupancy prediction results on the SurroundOcc-nuScenes validation set \cite{surroundocc}.} Our framework achieves state-of-the-art performance by a large margin with a sixfold improvement in FPS. All methods are benchmarked on the 4090.
     *GaussianWorld's paper results over-weight intermediate frames during evaluation. We re-evaluate released checkpoints under the standard setting. 
    }\label{tab:nuscenes-results}
\end{table*}

\begin{table*}[t] %
    \centering
    \vspace{-3mm}
    \setlength{\tabcolsep}{0.005\linewidth}   
    \renewcommand\arraystretch{1.05}
    \resizebox{1\linewidth}{!}{
    \begin{tabular}{l|c|c|c| c c c c c c c c c c c c c c c c c c}
    \toprule
    Method 
    & {\rotatebox{90}{Input}}  
    & IoU
    & mIoU
    &\rotatebox{90}{\textcolor{carcolor}{$\blacksquare$} car}
    &\rotatebox{90}{\textcolor{bicyclecolor}{$\blacksquare$} {bicycle}}
    &\rotatebox{90}{\textcolor{motorcyclecolor}{$\blacksquare$} {motorcycle}}
    &\rotatebox{90}{\textcolor{truckcolor}{$\blacksquare$} {truck}}
    &\rotatebox{90}{\textcolor{othervehiclecolor}{$\blacksquare$} {other-veh.}}
    &\rotatebox{90}{\textcolor{personcolor}{$\blacksquare$} {person}}
    &\rotatebox{90}{\textcolor{roadcolor}{$\blacksquare$} {road}}  
    &\rotatebox{90}{\textcolor{parkingcolor}{$\blacksquare$} {parking}}
    &\rotatebox{90}{\textcolor{sidewalkcolor}{$\blacksquare$} {sidewalk}}
    &\rotatebox{90}{\textcolor{othergroundcolor}{$\blacksquare$} {other-grnd}}
    &\rotatebox{90}{\textcolor{buildingcolor}{$\blacksquare$} {building}}
    &\rotatebox{90}{\textcolor{fencecolor}{$\blacksquare$} {fence}}
    &\rotatebox{90}{\textcolor{vegetationcolor}{$\blacksquare$} {vegetation}}
    &\rotatebox{90}{\textcolor{terraincolor}{$\blacksquare$} {terrain}}
    &\rotatebox{90}{\textcolor{polecolor}{$\blacksquare$} {pole}}
    &\rotatebox{90}{\textcolor{trafficsigncolor}{$\blacksquare$} {traf.-sign}}
    &\rotatebox{90}{\textcolor{other-struct.color}{$\blacksquare$} {other-struct.}}
    &\rotatebox{90}{\textcolor{other-objectcolor}{$\blacksquare$} {other-object}}
     
    \\\midrule

    LMSCNet~\cite{roldao2020lmscnet} & L & {47.53} & {13.65} & {20.91} & {0} & {0} & {0.26} & {0} & {0} & {62.95} & {13.51} & {33.51} & {0.2} & {43.67} & {0.33} & {40.01} & {26.80} & {0} & {0} & {3.63} & {0}
        
    \\ SSCNet~\cite{song2017semantic} & L & {53.58} & {16.95} & {31.95} & {0} & {0.17} & {10.29} & {0.58} & {0.07} & {65.7} & {17.33} & {41.24} & {3.22} & {44.41} & {6.77} & {43.72} & {28.87} & {0.78} & {0.75} & {8.60} & {0.67} 
    
    \\\midrule MonoScene~\cite{cao2022monoscene} & C & {37.87} & {12.31} & {19.34} & {0.43} & {0.58} & {8.02} & {2.03} & {0.86} & {48.35} & {11.38} & {28.13} & {3.22} & {32.89} & {3.53} & {26.15} & {16.75} & {6.92} & {5.67} & {4.20} & {3.09}
    
    \\ Voxformer~\cite{li2023voxformer} & C & {38.76} & {11.91} & {17.84} & {1.16} & {0.89}& {4.56} & {2.06}  & {1.63} & {47.01} & {9.67} & {27.21} & {2.89} & {31.18} & {4.97} & {28.99} & {14.69} & {6.51} & {6.92} & {3.79} & {2.43}
    
    \\ TPVFormer~\cite{huang2023tri} & C & {40.22} & {13.64} & {21.56} & {1.09} & {1.37} & {8.06} & {2.57} & {2.38} & {52.99} & {11.99} & {31.07} & {3.78} & {34.83} & {4.80} & {30.08} & {17.51} & {7.46} & {5.86} & {5.48} & {2.70} 
    
    \\ OccFormer~\cite{zhang2023occformer} & C & {40.27} & {13.81} & {22.58} & {0.66} & {0.26} & {9.89} & {3.82} & {2.77} & {54.30} & {13.44} & {31.53} & {3.55} & {36.42} & {4.80} & {31.00} & {19.51} & {7.77} & {8.51} & {6.95} & {4.60}

    \\ GaussianFormer~\cite{huang2024gaussianformer} & C & 35.38 & {12.92} & 18.93 & {1.02} & {4.62} & {18.07} & {7.59} & {3.35} & 45.47 & 10.89 & 25.03 & {5.32} & 28.44 & {5.68} & {29.54} & 8.62 & 2.99 & 2.32 & {9.51} & {5.14} \\
    
    GaussianFormer-2~\cite{huang2024probabilistic} & C & 38.37 & {13.90} & 21.08 & {2.55} & 4.21 & 12.41 & 5.73 & 1.59 & 54.12 & 11.04 & {32.31} & 3.34 & 32.01 & 4.98 & 28.94 & 17.33 & 3.57 & 5.48 & 5.88 & 3.54
 
\\
\midrule
 S2GO-Base (ours) & C  & \textbf{40.80} & \textbf{15.05}  & 22.72 & 1.28 & 1.66 & 15.87 & 5.13 & 2.07 & 53.77 & 13.31
 & 33.40 & 3.83  & 35.30 & 7.17 & 31.20 & 21.11 & 6.36 & 6.54 & 6.03 & 4.22 \\
\bottomrule
\end{tabular}
}
\vspace{-.5em}
\caption{\textbf{Results on the SSCBench-KITTI-360 test set \cite{kitti} with a monocular camera.}
\ourmethod achieves new state-of-the-art, achieving strong performance in all categories.
}
\label{tab:kitti-results}
\vspace{-.5em}
\end{table*}

\subsection{Stage 2: 3D Semantic Occupancy Prediction}\label{method:stage2}
\subsubsection{Occupancy Prediction Framework}
Equipped with the pre-training prior, \ourmethod is then trained for 3D semantic occupancy prediction. The model processes image observations, predicts a set of Gaussians $\mathcal{G}_t$ at each timestep, which now also include semantic class predictions, and ``splats'' Gaussians to nearby voxels as in Section \ref{method:prelim_gauss}. Notably, unlike the pre-training phase, query positions are initialized at learnable 3D locations. As such, our \ourmethod only uses RGB images during inference. The ``splatted'' voxel predictions are trained using ground truth, and we additionally supervise neighboring frames similar to Stage 1. In this section, we present strategies to further strengthen this pipeline.

\subsubsection{Opacity-Weighted Geometry Estimation}
Although the Gaussian-to-voxel splatting framework presented by GaussianFormer-2 elegantly handles foreground classes as a mixture of Gaussians, it only uses predicted opacity to weight Gaussians inside the mixture. As such, opacity has no bearing on determining binary occupancy of a location, in contrast to Gaussians in rendering \cite{kerbl20233d} where opacity acts as a proxy for density. This leads to unexpected behavior: Gaussians in background regions end up decreasing their scale $\textbf{s}$ and positioning themselves \textit{between} voxel centers to minimize their foreground contribution (Eq. \ref{eq:gaussian_prob}). This unnatural representation for Gaussians conflicts with the rendering initialization and hurts performance. To address this issue, we additionally weight the occupancy probability $\alpha(\mathbf{x}; \mathbf{G})$ with the opacity prediction $a$, yielding:
\begin{equation}\label{eq:gaussian_prob_updated}
    \alpha(\mathbf{x};\mathbf{G}) = a~ {\rm{exp}}\big(-\frac{1}{2}(\mathbf{x}-\mathbf{m})^{\rm T} \mathbf{\Sigma}^{-1} (\mathbf{x}-\mathbf{m})\big)
\end{equation}
This formulation improves foreground-background separation by allowing Gaussians in the background to simply predict lower opacity and by stabilizing the scale supervision to be more consistent between foreground and background regions.

\subsubsection{Efficient Gaussian-to-Voxel Splatting}
In Gaussian-to-voxel splatting, GaussianFormer \cite{gaussianformer} first determines pairs of interacting Gaussians and voxels, then parallelizes over voxels in the forward pass and over Gaussians in the backward pass. However, this formulation does not account for the inherent locality of the splatting operation --- neighboring voxels process a similar set of Gaussians and vice-versa. Such voxels and Gaussians should be processed together in a CUDA block for optimized L1 cache usage. This is especially a problem for the backward pass since naively parallelizing over Gaussians incurs random access costs on a large number of voxels (640k).

To address this problem in the forward pass, we block voxels into 4x4x4 grids and have threads tied to each voxel collaboratively load nearby Gaussians onto memory before splatting them, similar to 3DGS \cite{kerbl20233d}. In the backward pass, we adopt a similar approach but additionally take care to tie threads to individual Gaussians to avoid atomic operations on the gradients \cite{mallick2024taming}. Our efficient Gaussian-to-voxel splatting implementation, with 9k Gaussians and 640k voxels, speeds up the forward pass by \textbf{1.5x} (1.29ms to \textbf{0.87ms}) and the backward pass by \textbf{20.4x} (116ms to \textbf{5.7ms}), substantially reducing the wall-clock time required for training.

\subsubsection{Query Propagation}
A key point in our streaming 3D occupancy pipeline is query propagation. More specifically, we need to determine the optimal subset of current queries to push onto the queue for future timesteps. While a straightforward selection of top-k largest query opacities works well, maintaining the most occupied regions of the scene, we find that queries end up highly overlapping over time, with insufficient coverage over the scene. To mitigate this, we choose the highest opacity queries that are pairwise separated by a distance $\delta$, where $\delta$ is a hyperparameter. We find that this maintains an effective balance between maintaining high-opacity regions and distributing queries across the scene.

\section{Experiments}

\begin{table}[t]
  \setlength{\tabcolsep}{2pt}
  \centering
  \resizebox{\columnwidth}{!}{%
  \begin{tabular}{l|c|c|c|c|c c}
    \toprule
    Method & Backbone & Mask & RayIoU & mIoU & FPS \\
    \midrule
    BEVFormer~\cite{bevformer} & R101 & \checkmark & 32.4 & 39.2 & 3.0 \\
    RenderOcc~\cite{pan2024renderocc} & Swin-B & \checkmark & 19.5 & 24.4 & - \\
    SimpleOcc~\cite{gan2024comprehensive} & R101 & \checkmark & 22.5 & 31.8 & 9.7 \\
    BEVDet-Occ~\cite{huang2022bevdet4d} & R50 & \checkmark & 29.6 & 36.1 & 2.6 \\
    BEVDet-Occ-Long~\cite{huang2022bevdet4d} & R50 & \checkmark & 32.6 & 39.3 & 0.8 \\
    FB-Occ~\cite{li2023fb} & R50 & \checkmark & 33.5 & 39.1 & 10.3 \\
    \midrule
    BEVFormer~\cite{bevformer} & R101 & \xmark & 33.7 & 23.7 & 3.0 \\
    FB-Occ~\cite{liu2024fully} & R50 & \xmark & 35.6 & 27.9 & 10.3 \\
    SparseOcc~\cite{liu2024fully} & R50 & \xmark & 36.1 & 30.9 & 12.5 \\
    \midrule
    S2GO-Small (ours) & R50 & \xmark & 37.2 & 30.8 & \textbf{20.8} \\
    S2GO-Base (ours) & R50 & \xmark & \textbf{39.1} & \textbf{31.2} & 14.5 \\
    \bottomrule
  \end{tabular}}
  \vspace{-1em}
  \caption{\textbf{3D occupancy performance on Occ3D-nuScenes.}~\cite{tian2023occ3d}. We outperform prior work while maintaining a high FPS. FPS is measured on an A100.}
  \label{table:occ3d-nus}
  \vspace{-1em}
\end{table}

We perform extensive experiments on three benchmarks derived from the nuScenes and KITTI datasets. \ourmethod uses the ResNet50 \cite{he2016resnet} backbone, \ourmethod-Small uses 900 queries with 10 Gaussians each, and \ourmethod-Base uses 1800 queries with 20 Gaussians. Details about the datasets, metrics and the experiment setup can be found in Supplementary~\ref{supp:implementation}.

\subsection{Quantitative Results}
We first evaluate \ourmethod on the nuScenes dataset, with results provided in Table~\ref{tab:nuscenes-results}. On the SurroundOcc benchmark, \ourmethod-Small surpasses previous state-of-the-art GaussianWorld~\cite{zuo2024gaussianworld} by 1.50 IoU while offering a 5x increase in inference speed. Moreover, \ourmethod-Base further improves IoU by 1.19 and retains a 3x speed advantage. As shown in Table~\ref{table:occ3d-nus}, \ourmethod also achieves strong performance on the Occ3D benchmark, outperforming the fully sparse voxel-based method SparseOcc with fewer training epochs.

In addition to the nuScenes dataset~\cite{caesar2020nuscenes}, we also evaluate our approach on the KITTI-360 dataset~\cite{kitti}, with results summarized in Table~\ref{tab:kitti-results}. In this monocular 3D semantic occupancy prediction setting, \ourmethod again achieves state-of-the-art performance, surpassing GaussianFormer-2~\cite{huang2024probabilistic} by 8\% in mIoU and 6\% in IoU.

\begin{figure*}[t]
  \centering
  \vspace{-0.2in}
  \includegraphics[width=\linewidth]{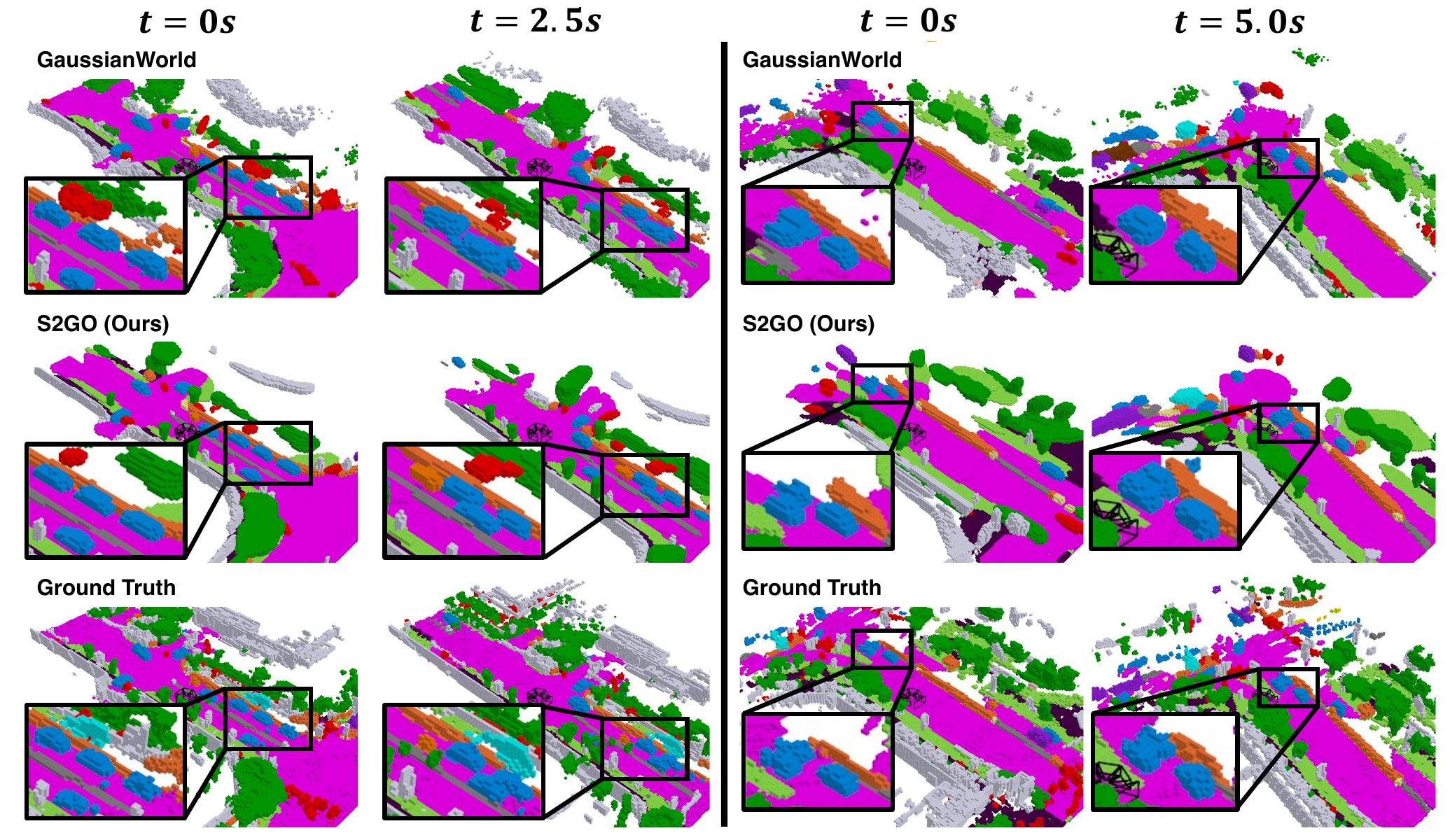}
  \vspace{-0.2in}
\caption{\textbf{Qualitative comparison of occupancy prediction.} We compare \ourmethod with GaussianWorld~\cite{zuo2024gaussianworld} by visualizing two timesteps from two distinct driving sequences. GaussianWorld struggles to maintain separate object representations over time, while \ourmethod effectively preserves distinct object identities by operating at a higher semantic level with sparse queries.}
  \label{fig:qual_surroundocc}
  \vspace{-1em}
\end{figure*}

\subsection{Qualitative Analysis} 
In Fig. \ref{fig:qual_surroundocc}, we present a qualitative comparison between our approach and GaussianWorld, visualizing two timesteps from two distinct driving sequences. Both methods successfully model individual vehicles in the initial frames. However, after several timesteps, when both the ego vehicle and surrounding vehicles have moved, GaussianWorld struggles to maintain independent representations of distinct objects and incorrectly merges multiple instances into one. This limitation arises because GaussianWorld, despite its streaming nature, directly operates on low-level Gaussian representations. Consequently, due to its weaker sense of objectness, local convolutions merge nearby objects. In contrast, \ourmethod{} decomposes the scene into a sparse set of queries, enabling it to operate at a higher semantic level and effectively preserve distinct object identities.

To demonstrate the capability of \ourmethod to model the dynamics of the driving world, we also visualize the future occupancy predictions in Fig. \ref{fig:future_prediction}.
\subsection{Ablations}

In this section, we verify the effectiveness of our proposed components. By default, models are trained for 12 epochs during both pretraining and occupancy prediction. All ablations are on the SurroundOcc-nuScenes dataset.

\begin{table}[t]
   \setlength{\tabcolsep}{5pt}
   \centering
   \resizebox{\columnwidth}{!}{
   \begin{tabular}{c|c|c|c|c|c|c}
      \toprule
      & Query Init. & Depth & RGB & Denoise & mIoU & IoU \\
      \midrule
      {(a)}  & - & \xmark & \xmark & \xmark & 13.02 & 25.73 \\
      {(a)†} & - & \xmark & \xmark & \xmark & 15.83 & 28.35 \\
      \midrule
      {(b)}  & Learnable  & \cmark & \cmark & \xmark & 12.42 & 26.64 \\
      {(c)}  & LiDAR  & \cmark & \cmark & \xmark & 13.62 & 27.08 \\
      {(d)}  & LiDAR+$\epsilon$ & \cmark & \cmark & \xmark & 20.55 & 32.68 \\
      \midrule
      {(e)}  & LiDAR+$\epsilon$ & \cmark & \xmark & \xmark & 20.25 & 32.44 \\
      {(d)}  & LiDAR+$\epsilon$ & \cmark & \cmark & \xmark & 20.55 & 32.68 \\
      {(f)}  & LiDAR+$\epsilon$ & \cmark & \cmark & \cmark & 21.60 & 33.91 \\
      \bottomrule
   \end{tabular}
   }
    \caption{\textbf{Ablation study on pretraining strategies.} LiDAR + $\epsilon$ denotes initialization from noised LiDAR. We find that pretraining with all objective is essential for occupancy prediction.}
   \label{table:ablation-init}
   \vspace{-1em}
\end{table}

\noindent \textbf{Pretraining.}
In Table \ref{table:ablation-init}, we ablate the impact of pretraining on \ourmethod and its formulations. Training occupancy prediction from scratch (a) yields inferior results, and training for 24 epochs (a)† only slightly improves performance. These results demonstrate that direct semantic occupancy training is insufficient due to ambiguous supervision.

We then include pretraining with depth and RGB supervision and ablate query position initialization. Learnable initialization -- which is what \ourmethod uses in the second stage -- is \textit{worse} than not pretraining. This occurs because the queries are randomly distributed throughout the 3D space, resulting in most queries being distant from any occupied geometry and therefore lacking adequate supervision. On the other hand, initializing query locations precisely at LiDAR points is only slightly better than not pretraining -- this baseline supervises Gaussians to capture local geometry, but the queries themselves are not supervised to move. Next, adding noise to LiDAR before initializing achieves remarkable performance, providing meaningful supervision to both queries and Gaussians. We emphasize that this is the only initialization method that substantially improves over not pretraining with the same compute budget (24 epochs of occupancy by (a)† vs 12+12 epochs with pretraining).

Finally, we ablate each pretraining loss function. Depth supervision alone is enough to achieve good performance. Adding RGB loss slightly boosts results as RGB supervises finer details, and denoising supervision gives a substantial final boost.

\begin{table}[t]
   \setlength{\tabcolsep}{5pt}
   \centering
   \resizebox{\columnwidth}{!}{
   \begin{tabular}{c|c|c|c|c}
      \toprule
      Opacity in $\alpha$ & Efficient G2V & mIoU & IoU & GPU hours \\
      \midrule
      \xmark & \xmark & 16.97 & 28.75 & 45h \\
      \cmark & \xmark & 20.13 & 32.28 & 93h \\
      \cmark & \cmark & 20.55 & 32.68 & 24h \\
      \bottomrule
   \end{tabular}
   }
   \caption{\textbf{Ablation on Gaussian-to-Voxel Splatting (G2V).} GPU hours are calculated for training 12 epochs on one A100. }
   \label{table:ablation-g2v}
\end{table}

\noindent \textbf{Gaussian to Voxel Splatting.}
In Table \ref{table:ablation-g2v}, we ablate our inclusion of opacity $a$ in occupancy probability $\alpha$ and our efficient Gaussian-to-voxel splatting implementation. First, excluding opacity substantially hurts geometry estimation of the model. Adding opacity estimation substantially improves performance (+3.16 mIoU), but doubles the training time as Gaussians opt to reduce occupancy probability by lowering opacity instead of scale, thus increasing the number of voxels each Gaussian affects. Leveraging our optimized CUDA kernels slightly improves performance while substantially lowering training costs, even in comparison to the original formulation without opacity in $\alpha$.

\begin{table}[t]
    \vspace{-.5em}
   \setlength{\tabcolsep}{5pt}
   \centering
   \scalebox{1.0}{
   \begin{tabular}{c|c|c}
      \toprule
      Propagation Type & mIoU & IoU \\
      \midrule
      None & 17.92 & 29.24 \\
      top-k opacity & 19.94 & 32.03 \\
      $\delta$-dist top-k opacity & 20.51 & 32.51 \\
      \bottomrule
   \end{tabular}
   }
    \caption{\textbf{Ablation on query propagation strategies.} "None" indicates no temporal information is used.}
   \label{table:ablation-query-prop}
   \vspace{-.5em}
\end{table}

\noindent \textbf{Query Propagation.}
Query selection for future frames is critical for streaming perception. In Table \ref{table:ablation-query-prop}, we ablate different propagation strategies. Compared to the single-frame baseline without propagation, selecting top-k queries by opacity already provides a substantial performance gain. However, this leads to excessive overlap between queries over time, wasting capacity in the model. Enforcing a minimum distance between queries encourages a more diverse spatial distribution, further improving performance.

\begin{table}[t]
   \setlength{\tabcolsep}{5pt}
   \centering
   \resizebox{\columnwidth}{!}{
   \begin{tabular}{c|c|c|c|c|c}
      \toprule
      \# Query & \# Gauss. / Query & \# Gauss.  & mIoU & IoU & FPS \\
      \midrule
      900  & 10 & 9000  & 21.60 & 33.91 & 20.8 \\
      1260 & 14 & 17640 & 21.78 & 34.15 & 17.9 \\
      1800 & 20 & 36000 & 21.84 & 34.51 & 14.5 \\
      \bottomrule
   \end{tabular}
   }
   \caption{\textbf{Ablation on the number of queries and Gaussians.} FPS is measured on an A100 GPU.}
   \label{table:ablation-queries}
\end{table}

\noindent \textbf{Number of Gaussians.}
In Table \ref{table:ablation-queries} we ablate the \# of queries and Gaussians. We observe that even just 900 sparse queries and 10 Gaussians per query is enable to capture the overall scene and achieve a high mIoU with a real-time 20.8 FPS on an A100. With more queries and Gaussians, the performance steadily improves, but at the cost of longer runtime. 

\begin{table}[t]
   \setlength{\tabcolsep}{5pt}
   \centering
   \scalebox{0.85}{
   \begin{tabular}{c|c|c|c}
      \toprule
      Pretraining & Occupancy Prediction & mIoU & IoU \\
      \midrule
      \xmark & \xmark & 20.07 & 31.87 \\
      \xmark & \cmark & 20.15 & 31.94 \\
      \cmark & \xmark & 20.50 & 32.62 \\
      \cmark & \cmark & 20.55 & 32.68 \\
      \bottomrule
   \end{tabular}
   }
   \caption{\textbf{Ablation on using velocity modeling in each stage.}}
   \label{table:ablation-temporal}
\end{table}

\begin{figure}[t]
  \centering
  \includegraphics[width=\linewidth]{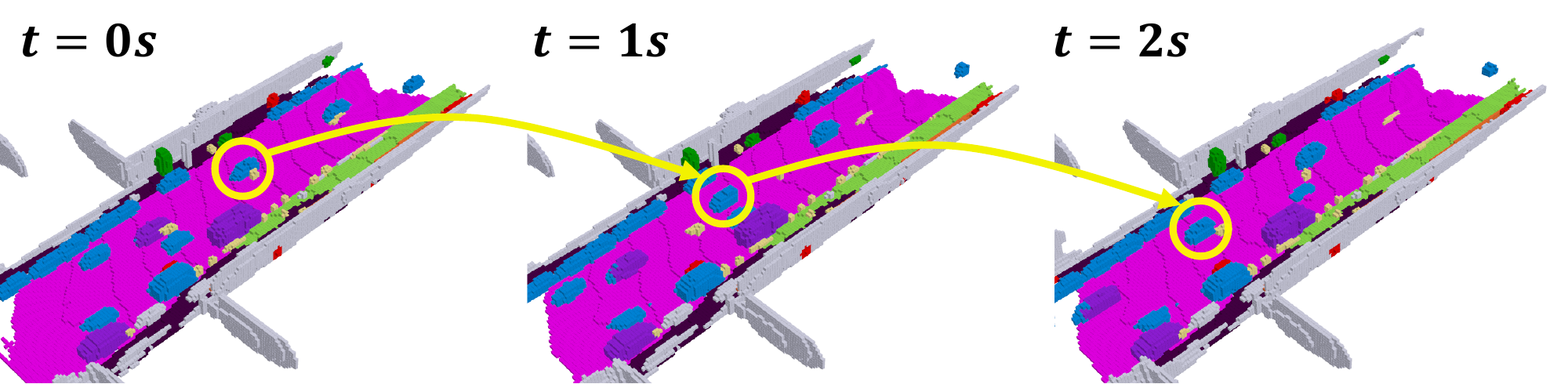}
  \vspace{-0.2in}
  \caption{\textbf{Visualization of future occupancy predictions.} We use the self-supervised velocity prediction for each query to roll out future occupancy predictions. Our streaming query-based framework well-decouples motion of individual objects.}
  \label{fig:future_prediction}
\end{figure}

\noindent \textbf{Velocity Modeling.}
\ourmethod predicts a velocity for each query, which is used in both stages to move dynamic regions before applying RGBD or occupancy supervision in neighboring frames. While this module is useful on its own for future occupancy prediction as shown in Fig. \ref{fig:future_prediction}, we ablate its impact on performance in Table \ref{table:ablation-temporal}. Velocity modeling improves performance in both stages, with motion modeling during pretraining proving particularly important.

\section{Conclusion and Future Work}
We presented a novel framework for 3D semantic occupancy prediction that leverages sparse 3D queries to efficiently capture and propagate scene information over time. Our method replaces traditional dense, grid-aligned Gaussian representations with a compact, streaming set of semantic queries. A geometry denoising pre-training phase ensures effective alignment of sparse queries with dense occupancy targets, accurately modeling both static and dynamic scene elements. Extensive evaluations on nuScenes and KITTI benchmarks demonstrate state-of-the-art performance while operating 5.9× faster than previous methods. Our work demonstrates that a query-based approach can effectively bridge the gap between efficiency and high-fidelity 3D scene representation. In the future, we plan to explore multitask, end-to-end learning and large-scale pretraining using unlabeled data to further enhance model performance and generalization.

\noindent\textbf{Acknowledgments.} We would like to thank Ryan Brigden for infrastructure support as well as Vickram Rajendran and Stephen Yang for paper writing help. 

{
    \small
    \bibliographystyle{ieeenat_fullname}
    \bibliography{main}
}

\clearpage
\setcounter{page}{1}

\maketitlesupplementary

\begin{figure*}[t]
  \centering
  \includegraphics[width=0.8\linewidth]{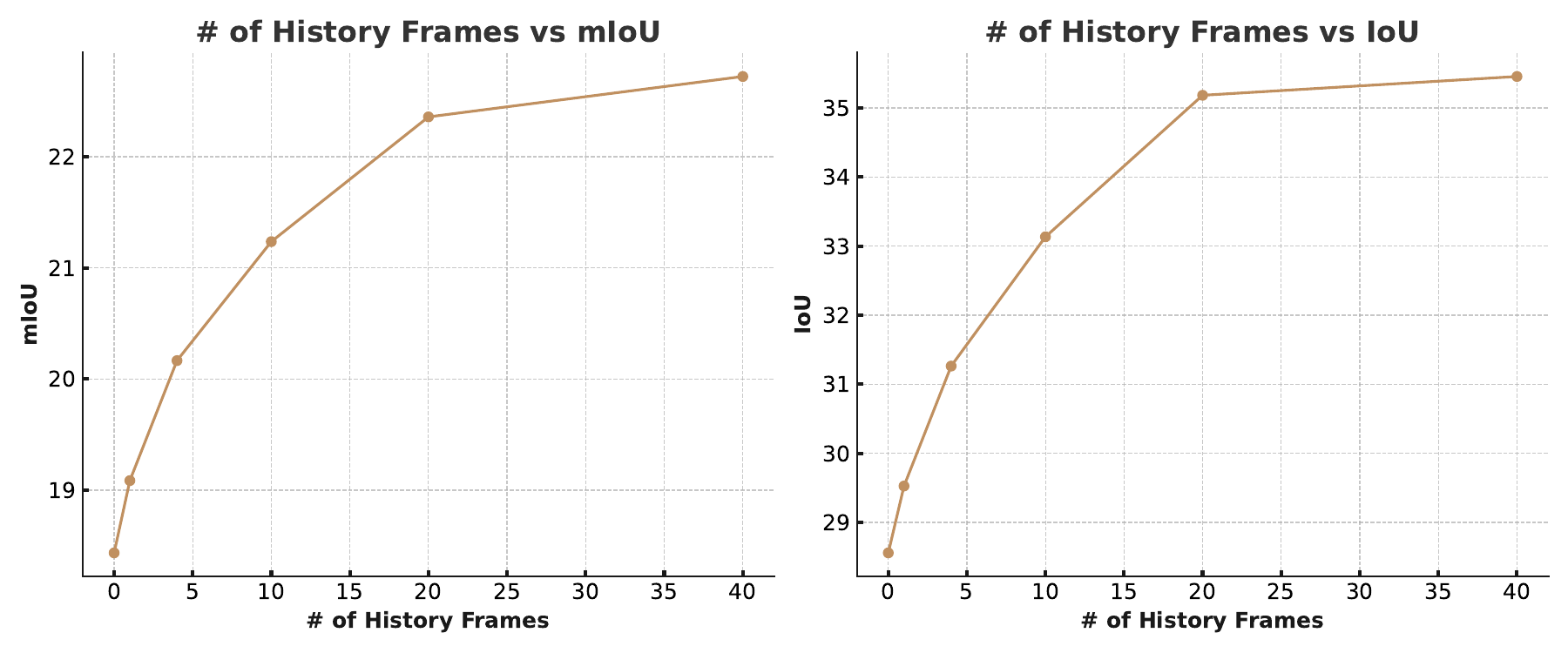}
    \caption{\textbf{Impact of history length on occupancy performance.} A longer history consistently improves performance, showcasing the advantage of our streaming approach over prior projection-based methods.}
  \label{fig:history_frames_no_5_vs_miou_iou}
\end{figure*}

\begin{table*}[t]
  \setlength{\tabcolsep}{5pt}
   \centering
   \vspace{-3pt}
   \scalebox{0.91}{
   \begin{tabular}{l|cccc|c|ccc|c|c}
      \toprule
      Method  & Backbone  &Mask& Input Size & Epoch & \cellcolor[gray]{0.93}{RayIoU} & \multicolumn{3}{c|}{RayIoU\textsubscript{1m, 2m, 4m}} & mIoU & FPS \\
      \midrule
      BEVFormer~\cite{bevformer}  & R101    & \checkmark & 1600$\times$900 & 24 & \cellcolor[gray]{0.93}{32.4} & 26.1 & 32.9 & 38.0 & 39.2 & 3.0 \\
      RenderOcc \cite{pan2024renderocc}  & Swin-B  & \checkmark & 1408$\times$512 & 12 & \cellcolor[gray]{0.93}{19.5} & 13.4 & 19.6 & 25.5 & 24.4 & - \\
      SimpleOcc \cite{gan2024comprehensive}  & R101  & \checkmark & 672$\times$336 & 12 & \cellcolor[gray]{0.93}{22.5} & 17.0 & 22.7 & 27.9 & 31.8 & 9.7 \\
      BEVDet-Occ~\cite{huang2022bevdet4d}  & R50     & \checkmark & 704$\times$256  & 90 & \cellcolor[gray]{0.93}{29.6} & 23.6 & 30.0 & 35.1 & 36.1 & 2.6 \\
      BEVDet-Occ-Long~\cite{huang2022bevdet4d}   & R50     & \checkmark & 704$\times$384  & 90 & \cellcolor[gray]{0.93}{32.6} & 26.6 & 33.1 & 38.2 & {39.3} & 0.8 \\
      FB-Occ~\cite{li2023fb}        & R50     & \checkmark & 704$\times$256  & 90 & \cellcolor[gray]{0.93}{33.5} & 26.7 & 34.1 & 39.7 & 39.1 & 10.3 \\
      
      \midrule
      BEVFormer~\cite{bevformer} & R101 & \xmark & 1600$\times$900 & 24 & \cellcolor[gray]{0.93}{{33.7}} & - & - & - & 23.7 & 3.0 \\
      FB-Occ~\cite{liu2024fully} & R50 & \xmark & 704$\times$256 & 90 & \cellcolor[gray]{0.93}{{35.6}} & - & - & - & 27.9 & 10.3 \\
      SparseOcc~\cite{liu2024fully} & R50 & \xmark & 704$\times$256 & 48 & \cellcolor[gray]{0.93}{{36.1}} & {30.2} & {36.8} & {41.2} & 30.9 & 12.5 \\
      \midrule
      S2GO-Small (ours)  & R50 & \xmark & 704$\times$256 & 24 &  \cellcolor[gray]{0.93}{37.2} & 31.3 & 38.1 & 42.2 & 30.8 & 20.8 \\
      S2GO-Base (ours)  & R50 & \xmark & 704$\times$256 & 24 &  \cellcolor[gray]{0.93}{\textbf{39.1}} & 33.1 & 40.0 & 44.1 & 31.2 & 14.5 \\
      \bottomrule
   \end{tabular}
   }
  \caption{\textbf{3D occupancy prediction performance on the Occ3D-nuScenes validation set} \cite{tian2023occ3d}. }
\label{table:occ3d-nus-full}
\end{table*}

\section{Experiment setup}
\label{supp:implementation}
\noindent \textbf{Datasets.}
We conducted comprehensive experiments on three benchmarks derived from nuScenes and KITTI.
The \textit{nuScenes dataset}~\cite{caesar2020nuscenes} provides 1000 scenes of surround-view driving scenes. We evaluate our method on both the SurroundOcc~\cite{surroundocc} and Occ3D~\cite{tian2023occ3d} benchmarks. SurroundOcc provides voxel-based annotations in a 100 × 100 × 8 m² range around the car with a 200 × 200 × 16 resolution, classifying voxels into 18 classes (16 semantic, 1 empty, and 1 noise). Occ3D offers voxelized semantic occupancy in a 80 × 80 × 6.4 m² range with a 200 × 200 × 16 resolution, derived from an auto-labeling pipeline.
The \textit{KITTI dataset}~\cite{kitti} comprises over 320k images and 80k laser scans from suburban driving scenes. We adopt the dense semantic annotations from SSCBench-KITTI-360~\cite{li2024sscbench,liao2022kitti}. The official split consists of 7/1/1 sequences for training, validation, and testing, respectively. The voxel grid spans an area of 51.2 × 51.2 × 6.4 m² in front of the ego car, with a resolution of 256 × 256 × 32. Each voxel is classified into one of 19 classes (18 semantic categories and 1 empty).

\noindent \textbf{Evaluation Metrics.}
Following MonoScene~\cite{cao2022monoscene}, we use \textbf{IoU}and \textbf{mIoU} as evaluation metrics. For the Occ3D dataset, we adopt RayIoU as our primary metric following SparseOcc~\cite{liu2024fully}, \textbf{RayIoU} extends mIoU by evaluating occupancy predictions at the ray level rather than voxel level. It simulates LiDAR rays and assesses predictions based on both depth accuracy and class correctness. RayIoU ensures balanced evaluation by resampling rays across distances and incorporating temporal casting from past, present, or future viewpoints to assess scene completion. By preventing inflated IoU scores caused by thick surface predictions and applying a depth threshold for true positive classification, RayIoU provides a more robust evaluation. Metrics are defined as:
\vspace{-8pt}
\begin{equation}
  \text{mIoU/RayIoU} = \frac{1}{|C|} \sum_{i \in C} \frac{TP_i}{TP_i + FP_i + FN_i}
\end{equation}
\vspace{-10pt}
\begin{equation}
\text{IoU} = \frac{TP_{\neq c_0}}{TP_{\neq c_0} + FP_{\neq c_0} + FN_{\neq c_0}}
\end{equation}

where $TP_i$, $FP_i$, and $FN_i$ are the number of true positive, false positive, and false negative predictions for class $i$, $C$ is the set of semantic classes, and $c_0$ is the nonempty class. For RayIoU, a query ray is classified as a true positive (TP) if the predicted class matches the ground truth and the L1 error between the predicted and ground-truth depth is within a certain threshold (1m, 2m, 4m)

\noindent \textbf{Baselines}
We evaluate \ourmethod against representative approaches spanning diverse 3D representation paradigms. Specifically, we compare with voxel-based methods, including MonoScene~\cite{monoscene}, Atlas~\cite{murez2020atlas}, SurroundOcc~\cite{surroundocc}, which employ dense 3D voxel grids for occupancy reconstruction. We further benchmark against BEV-based methods like BEVFormer~\cite{bevformer}. In addition, we consider the triplane-based TPVFormer~\cite{tpvformer}, which decomposes 3D space into orthogonal 2D planes, facilitating efficient feature aggregation. Lastly, we include Gaussian-based approaches—GaussianFormer~\cite{gaussianformer}, GaussianFormer-2~\cite{huang2024probabilistic}, and GaussianWorld~\cite{zuo2024gaussianworld}—which employ 3D Gaussians to model 3D occupancy and semantics.

\begin{figure*}[t]
  \centering
  \vspace{-0.2in}
  \includegraphics[width=\linewidth]{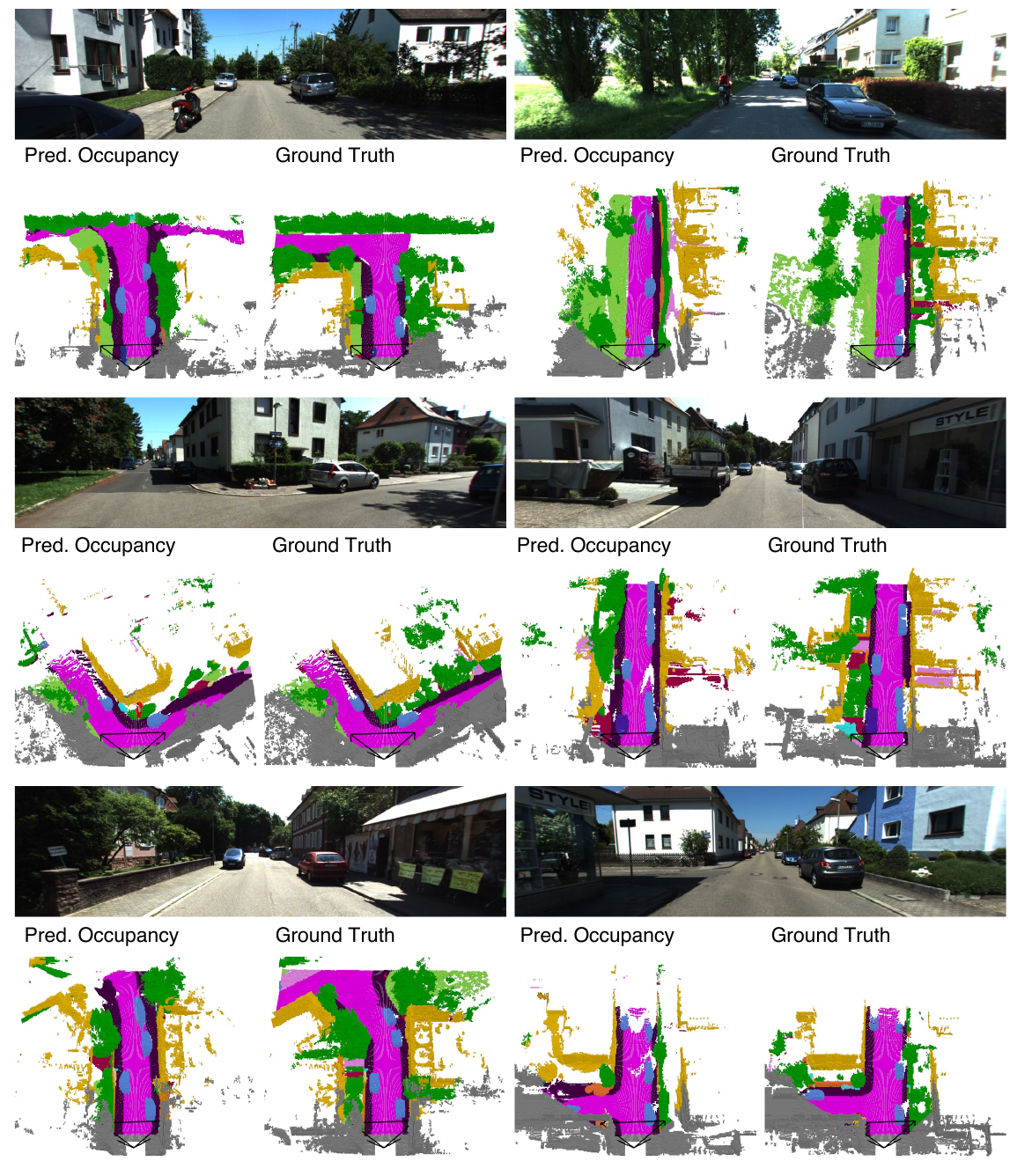}
  \vspace{-0.2in}
\caption{\textbf{Qualitative Results on the SSCBench-KITTI-360 dataset.} \ourmethod well-captures occupancy details even in a monocular setting.}
  \label{fig:kitti}
\end{figure*}

\section{Implementation details}
On nuScenes, \ourmethod uses a 256x704 resolution image and is pre-trained on denoising and rendering for 12 epochs without semantic annotations, and then trained for 24 epochs for 3D semantic occupancy prediction. \ourmethod-Small uses an ImageNet1k backbone, while \ourmethod-Base leverages nuImages pre-training. On KITTI, we use a 256x1408 resolution image and an ImageNet1k backbone. The model is pre-trained for 12 epochs, then trained for occupancy for another 12 epochs.

The temporal transformer closely follows the design from PETR \cite{petr} and StreamPETR \cite{streampetr}, with a 4-frame (2s) queue. All models are trained with a 4e-4 learning rate with a batch size of 16, with the cosine annealing schedule. On nuScenes-SurroundOcc, the LiDAR nosing factor $\epsilon$ is set to 1 meter. During training, the pairwise query distance $\delta$ for query propagation is randomly sampled between 0 to 3 meters, and during inference, it is set to 1.6m. For nuScenes-Occ3D and KITTI, all distances are scaled according to the smaller extent of the 3D scene. The embedding dimension of the temporal transformer is 768, and we leverage Flash Attention \cite{dao2022flashattention} for efficient self-attention between queries. Queries interact with the image through Deformable Attention \cite{zhu2020deformable,sparse4d,streampetr}.

\section{Number of History Frames}
To further evaluate \ourmethod, we plot occupancy performance over different streaming history lengths in Figure \ref{fig:history_frames_no_5_vs_miou_iou}. With a longer history, performance steadily improves, demonstrating the efficacy of our streaming framework. We emphasize that unlike prior projection-based works, \ourmethod incurs \textit{no additional cost} from a longer history.

\section{Latency Breakdown}
We benchmark our 9000 Gaussian model on an A100 GPU. The backbone, temporal transformer, gaussian prediction, and propagation take 11.54ms, 22.79ms, 2.22ms, and 1.45ms, respectively.

\begin{table}[t]
\vspace{-1.5em}
\centering
\renewcommand{\arraystretch}{0.8}
\small
\begin{tabular}{lcc}
\toprule
\textbf{Pretraining Query Init.} & \textbf{mIoU} & \textbf{IoU} \\
\midrule
LiDAR & 21.60 & 33.91 \\
Zero-shot RGB depth estimation \cite{yin2023metric} & 20.99 & 33.57 \\
\bottomrule
\end{tabular}
\vspace{-1em}
\caption{Ablation of different pretraining query initializations.}
\label{pretrain_ablation}
\end{table}

\section{Pre-training with Zero-shot Monocular Depth}
In Table \ref{pretrain_ablation} we ablate the use of LiDAR during pre-training by replacing it with zero-shot monocular depth predictions from Metric3D \cite{yin2023metric} on RGB images. We find that this largely maintains performance, indicating the generality of our pretraining pipeline.

\section{Comprehensive Evaluation results}
We provide extensive comparisons with existing methods on the Occ3D benchmark using detailed metrics, as shown in Tab.~\ref{table:occ3d-nus-full}.

\section{More Qualitative Results}
In Fig. \ref{fig:kitti} we visualize example predictions and ground truth from the SSCBench-KITTI-360 dataset. Our framework flexible adapts to a monocular setting and precisely predicts the semantic occupancy of the driving scene.

\end{document}